\documentclass[12pt,letterpaper]{article}
\usepackage{osajnl2}
\usepackage{paralist}
\usepackage{amsmath}
\usepackage{amsfonts}
\usepackage{subfigure}
\usepackage{multirow}
\usepackage{fancyhdr}

\newtheorem{theorem}{Theorem}[section]

\begin{document}

\title{Cost-Effective Implementation of Order-Statistics Based Vector Filters Using Minimax Approximations}

\author{M. Emre Celebi$^{1,*}$, Hassan A. Kingravi$^2$, Rastislav Lukac$^{3}$, and Fatih Celiker$^4$}
\address{$^1$Dept.\ of Computer Science, Louisiana State University, Shreveport, LA, USA}
\address{$^2$Dept.\ of Computer Science, Georgia Institute of Technology, Atlanta, GA, USA}
\address{$^3$Epson Edge, Epson Canada Ltd., Toronto, Ontario, Canada}
\address{$^4$Dept.\ of Mathematics, Wayne State University, Detroit, MI, USA}
\address{$^*$Corresponding author: ecelebi@lsus.edu}

\begin{abstract}

Vector operators based on robust order statistics have proved successful in digital multichannel imaging applications, particularly color image filtering and enhancement, in dealing with impulsive noise while preserving edges and fine image details. These operators often have very high computational requirements which limits their use in time-critical applications. This paper introduces techniques to speed up vector filters using the minimax approximation theory. Extensive experiments on a large and diverse set of color images show that proposed approximations achieve an excellent balance among ease of implementation, accuracy, and computational speed.

\end{abstract}


\maketitle

\section{Introduction}
\label{sec_intro}

Image noise filtering - the process of estimating the original image information from noisy data - is a common preprocessing step in image processing and analysis applications, as the presence of noise in images not only lowers their perceptual quality, but also makes subsequent tasks such as edge detection and segmentation more difficult \cite{Plataniotis00}. With the recent shift from traditional grayscale imaging to color imaging, numerous filters have been proposed for removing noise from color images. An extensive overview of color image filtering solutions and their applications can be found in \cite{Lukac06a}, with detailed performance analysis presented in \cite{Celebi07a}.
\par
An important class of filters for noise reduction in color images is the one based on robust vector order statistics \cite{Lukac05,Celebi08}. A typical natural image exhibits strong correlation between its red, green, and blue color channels; therefore, treating the pixels of the image as vectors avoids color shifts and artifacts in the output of the filter. Since images are nonstationary due to the presence of edges as well as noise and blur introduced during the image formation, vector filters usually operate on pixels inside a supporting window that slides over the image. Desired noise filtering characteristics can be obtained by using vectors with certain ranks in the ordered set of pixel values inside the supporting window, as an ordering operation performed according to a distance or similarity criterion distinguishes outliers from noise-free samples \cite{Lukac05}.
\par
Many researchers have noted the high computational requirements of order-statistics based vector filters; however, relatively few studies \cite{Barni97,Barni00} have focused on alleviating this problem. Furthermore, the scope of these studies is limited to the vector median filter \cite{Astola90}, which has been considered as the gold standard of performance in color image filtering due to its robustness and excellent impulsive noise suppression capability \cite{Celebi07b}.
\par
This paper introduces techniques to speed up popular vector filters which use vector ordering criteria other than the Euclidean distance. In particular, the filtering solutions from \cite{Trahanias93,Plataniotis97,Lukac03a} involve, respectively, computationally expensive inverse cosine, exponential, and logarithmic functions that are evaluated during the filtering process typically millions of times. In order to allow the use such filters in time-critical imaging applications, we utilize the minimax approximation theory to substitute the abovementioned elementary functions with computationally efficient polynomials. Extensive experiments on a large and diverse image set show that the presented approximations achieve an excellent balance among ease of implementation, accuracy, and computational speed.
\par
The rest of the paper is organized as follows. Section \ref{sec_minimax} gives background on minimax approximation theory. Section \ref{sec_impl} introduces the use of the minimax approximation theory in speeding up order-statistics based vector filters. Motivation and design characteristics are discussed in detail. Section \ref{sec_exp} describes the image set, noise models, filtering performance criteria, and the experimental setup. Finally, conclusions are given in Section \ref{sec_conc}.

\section{Overview of Minimax Approximation Theory}
\label{sec_minimax}

Given a function $f$, we would like to approximate it by another function $g$ so that the error ($\varepsilon$) between them over a given interval is arbitrarily small. The existence of such approximations is stated by the following theorem:

\begin{theorem}\emph{(Weierstrass)}
Let $f$ be a continuous real-valued function defined on $[a,b]$, i.e.\ $f \in C[a,b]$. Then $\forall \varepsilon > 0$ there exists a polynomial $P$ such that $\| f - P \| < \varepsilon$, i.e.\ $\forall z \in [a,b], \ \left| f(z)-P(z) \right| < \varepsilon$.
\end{theorem}

This is commonly known as the minimax approximation to a function. It differs from other methods, e.g.\ least squares approximations, in that it minimizes the maximum error ($\varepsilon$) rather than the average error:
\begin{equation}
\varepsilon = \mathop {\rm max}\limits_{z \in [a,b]} \left| f(z)-P(z) \right|.
\end{equation}
A similar theorem establishes the existence of a rational variant of this method \cite{Cheney00}. Let $n \geq 0$ be a natural number and let
\begin{equation}
P_n \left( [a,b] \right) = \left\{ {a_0  + a_1 z + \, \ldots \, + a_n z^n :z \in [a,b],\;\;a_i  \in \mathbb{R},\;\;i = 0, 1, \ldots, n} \right\}
\end{equation}
be the set of all polynomials of degree less than or equal to $n$. The set of irreducible rational functions,
$R^n _m \left( {[a,b]} \right)$, is defined as
\begin{equation}
R^n _m \left( {[a,b]} \right) = \left\{ {\frac{{p(z)}}
{{q(z)}}:\,p(z)\, \in P_n \left( {[a,b]} \right),\,\,q(z) \in P_m \left( {[a,b]} \right)} \right\}
\end{equation}
where $p$ and $q$ have no common factors. Then \cite{Cheney00}:
\begin{theorem}
For each function $f \in C[a,b]$, there exists at least one best rational approximation from the class $R^n _m \left( {[a,b]} \right)$.
\end{theorem}

This theorem states the existence of a rational approximation $r^* \in R^n_m \left( {[a,b]} \right)$ to a function $f \in C[a,b]$ that is optimal in the Chebyshev sense:
\begin{equation}
\mathop {\rm max}\limits_{z \in [a,b]} \left| f(z)-r^*(z) \right| = \mathop{\rm dist}\left( f,R^n_m \right)
\end{equation}

where $\mathop{\rm dist}\left( f,R^n_m \right)$ denotes the distance between $f$ and $R^n_m \left( {[a,b]} \right)$ with respect to some norm, in our case, the Chebyshev (maximum) norm. Regarding the choice between a polynomial and a rational approximant, it can be said that certain functions can be approximated more accurately by rationals than by polynomials. Jean-Michel Muller explains this phenomenon as follows ``It seems quite difficult to predict if a given function will be much better approximated by rational functions than by polynomials. It makes sense to think that functions that have a behavior that is 'highly nonpolynomial' (finite limits at $\pm \infty$, poles, infinite derivatives, $\ldots$) will be poorly approximated by polynomials.'' \cite{Muller06}.

In this study the Remez Exchange Algorithm, an iterative method that uses Lagrangian interpolation to systematically minimize the maximum absolute difference between the given function and its polynomial approximation, was used to calculate the polynomials. The reader is referred to \cite{Cheney00,Muller06} for more information on the minimax approximation theory and \cite{Fraser65} for the implementation details of the Remez algorithm.

\section{Proposed Implementations of Vector Filters}
\label{sec_impl}

\begin{figure}[!ht]
\centering
\includegraphics*[width=0.20\columnwidth,draft=false]{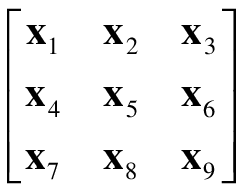}
\caption { \label{fig_window} Indexing convention inside a $3 \times 3$ window }
\end{figure}

Consider an $M \times N$ red-green-blue (RGB) input image {\bf X} that represents a two-dimensional array of three-component vectors ${\bf x}(r,c) = [x_1(r,c),x_2 (r,c),x_3 (r,c)]$ occupying the spatial location $(r,c)$, with the row and column indices $r = \left\{ 1, \ldots, M \right\}$ and $c = \left\{ 1, \ldots, N \right\}$,  respectively. In the pixel ${\bf x}(r,c)$, the $x_k(r,c)$ values denote the red ($k=1$), green ($k=2$), and blue ($k=3$) components. In order to isolate small image regions, each of which can be treated as stationary, and reduce processing errors by operating in such a localized area of the input image, an $\sqrt{n} \times \sqrt{n}$ supporting window $W(r,c)$ centered on pixel ${\bf x}(r,c)$ is used. The window slides over the entire image {\bf X} in a raster fashion and the procedure replaces the input vector ${\bf x}(r,c)$ with the output vector ${\bf y}(r,c) = F(W(r,c))$ of a filter function $F(\cdot)$ that operates over the samples inside $W(r,c)$. Repeating the procedure for each pair $(r,c)$, with $r = \left\{ 1, \ldots, M \right\}$ and $c = \left\{ 1, \ldots, N \right\}$, produces the output vector ${\bf y}(r,c)$ of the $M \times N$ filtered image {\bf Y}. For notational simplicity, the input vectors inside $W(r,c)$ are re-indexed as a set, i.e.\ $W(r,c) = \left\{ {\bf x}_i: i=1, \ldots, n \right\} $(see Figure \ref{fig_window}), as commonly seen in the related literature \cite{Lukac06a, Celebi07a}. In this notation, the center pixel in $W$ is given by ${\bf x}_C = {\bf x}_{(n+1)/2}$ and in the vector ${\bf x}_i = [x_{i1}, x_{i2}, x_{i3}]$ with components $x_{ik}$, the $i \in \{1, \ldots, n\}$ and $k \in \{1,2,3\}$ indices denote the position of the vector inside the window and the color channel, respectively.


\subsection{Vector Directional Filters}
\label{sec_vdf}

The vector directional filter (VDF) family \cite{Trahanias93} operates on the direction of the input vectors with the aim of eliminating the vectors with atypical directions. This family utilizes the angle between the input vectors to order the vectors inside the supporting window. For example, the output of the basic vector directional filter (BVDF), the most well-known member of the VDF class, is the input vector inside the supporting window whose direction is the maximum likelihood estimate of the directions of the input vectors \cite{Nikolaidis98}:
\begin{equation}
\begin{array}{l}
  \mathbf{y}(r,c) = \mathop {\rm argmin}\limits_{\mathbf{x}_i  \in W(r,c)} \left( {\sum\limits_{j = 1}^n {A(\mathbf{x}_i ,\mathbf{x}_j )} } \right) \hfill \\
  A\left( {\mathbf{x}_i, \mathbf{x}_j} \right) = \arccos \left( {\frac{{x_{i1} x_{j1}  + x_{i2} x_{j2}  + x_{i3} x_{j3} }}
{{{\parallel {\mathbf{x}_i} \parallel}_2 {\parallel {\mathbf{x}_j} \parallel}_2 }}} \right) \hfill \\
\end{array}
\end{equation}
where $A(\mathbf{x}_i, \mathbf{x}_j)$ denotes the angle between the two input vectors $\mathbf{x}_i$ and $\mathbf{x}_j$ and ${\parallel . \parallel}_2$ is the $L_2$ (Euclidean) norm. Note that in addition to BVDF, the angular function $A(.,.)$ was used in the design of a number of other filters including the generalized VDF \cite{Trahanias96}, directional distance filter \cite{Karakos97}, hybrid vector filters \cite{Khriji02}, weighted VDFs \cite{Lukac04}, data-adaptive VDFs \cite{Plataniotis98}, and switching VDFs \cite{Lukac06b}.
\par
\begin{figure}[!ht]
\centering
\includegraphics[width=0.5\columnwidth,draft=false]{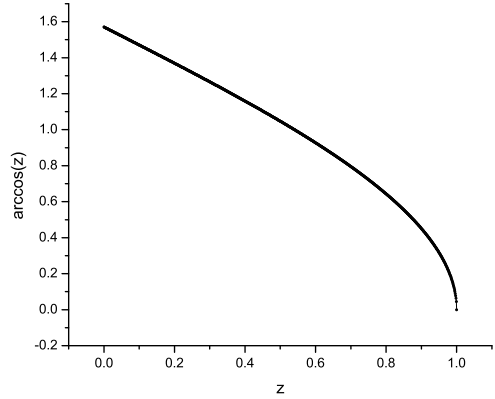}
\caption { \label{fig_arccos} Function arccos(z) in the interval $[0,1]$ }
\end{figure}
\begin{figure}[!ht]
\centering
\includegraphics[width=0.5\columnwidth,draft=false]{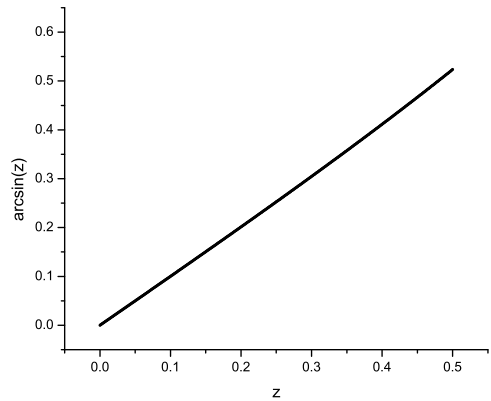}
\caption { \label{fig_arcsin} Function arcsin(z) in the interval $[0,0.5]$ }
\end{figure}

The computational requirements of these filters can be reduced by speeding up the inverse cosine (ARCCOS) function, whose argument falls into the interval $[0,1]$ (see Figure \ref{fig_arccos}). Unfortunately, approximating the ARCCOS function in this interval is not easy because of its behavior near $1$. This can be circumvented using the following numerically more stable identity for $z \geq 0.5$:
\begin{equation}
\arccos{(z)} = 2\arcsin \left( {\sqrt {0.5(1 - z)} } \right)
\end{equation}
where the inverse sine function (ARCSIN) receives its arguments from the interval $[0,0.5]$ (see Figure \ref{fig_arcsin}). Instead of plugging the value of $\sqrt {0.5(1 - z)}$ into a minimax approximation for the ARCSIN function and then multiplying the result by $2$,
two multiplication operations can be avoided if the following function is approximated:
\begin{equation}
\begin{array}{l}
  \tau  = \sqrt{1 - z} \hfill \\
  \arccos{(z)} = 2\arcsin{\left( {\tau /\sqrt 2 } \right)} \hfill \\
\end{array}
\end{equation}
where the argument $\tau$ falls into the interval
$\,\left[ {0,\,{1 \mathord{\left/ {\vphantom {1 {\sqrt 2 }}} \right.
 \kern-\nulldelimiterspace} {\sqrt 2 }}} \right]$.

Table \ref{tab_arcsin_arccos} lists the coefficients of the fourth degree minimax polynomials that approximate the ARCSIN and ARCCOS functions. Since both functions exhibit strong linearity in their respective intervals, they can be accurately approximated by polynomials, as indicated by the small error values listed in the table.

\begin{table}
\centering
\scriptsize
{
\caption{ \label{tab_arcsin_arccos} Fourth degree minimax polynomials for the ARCSIN and ARCCOS functions }
\begin{tabular}{ c|c|c|c|c|c|c }
\hline
Function & $\varepsilon$ & $a_0$ & $a_1$ & $a_2$ & $a_3$ & $a_4$\\
\hline
\hline
ARCSIN & 2.097814e-05 & 2.097797e-05 & 1.412840 & 1.429881e-02 & 6.704361e-02 & 6.909677e-02\\
\hline
ARCCOS & 1.048949e-05 & 1.570786 & -9.990285e-01 & -1.429899e-02 & -9.481335e-02 & -1.381942e-01\\
\hline
\end{tabular}
}
\end{table}

\subsection{Adaptive Multichannel Non-Parametric Filters}
\label{sec_amnf}

Adaptive Multichannel Non-Parametric Filters (AMNFs) \cite{Plataniotis97} approach the filtering problem from an estimation theoretic perspective. Specifically, these filters employ non-parametric kernel density estimators to determine the pixels in the filtered image as follows:
\begin{equation}
\label{equ_amnf}
\begin{array}{l}
  {\bf y}(r,c) = \sum\limits_{i = 1}^n {{\bf x}_i \left( {\frac{{h_i ^{ - 3} K\left( {{{\left( {{\bf x}_C  - {\bf x}_i } \right)} \mathord{\left/
 {\vphantom {{\left( {{\bf x}_C  - {\bf x}_i } \right)} {h_i }}} \right.
 \kern-\nulldelimiterspace} {h_i }}} \right)}}
{{\sum\nolimits_{j = 1}^n {h_j ^{ - 3} K\left( {{{\left( {{\bf x}_C - {\bf x}_j } \right)} \mathord{\left/
 {\vphantom {{\left( {{\bf x}_C - {\bf x}_j} \right)} {h_j }}} \right.
 \kern-\nulldelimiterspace} {h_j }}} \right)} }}} \right)}  \hfill \\
  h_i  = n^{ - \kappa /3} \sum\limits_{j = 1}^n {{\parallel {{\bf x}_i - {\bf x}_j} \parallel}_1 }  \hfill \\
\end{array}
\end{equation}
where ${\parallel . \parallel}_1$ denotes the $L_1$ (City-Block) norm. Two possible choices for the kernel function are the multivariate exponential $K(\mathbf{x}) = e^{-{\parallel {\bf x} \parallel}_1}$ (AMNFE) and the multivariate gaussian $K(\mathbf{x}) = e^{ - 0.5\, {\parallel {\bf x} \parallel}_2^2}$ (AMNFG) functions. The scaling factor $\kappa$ in the kernel width calculation is set to the author recommended value of 0.33 \cite{Plataniotis97}. The computational requirements of (\ref{equ_amnf}) can be reduced by speeding up the kernel computation. Both kernels involve the exponential (EXP) function which can be accurately approximated by polynomials. Note that in addition to AMNFs, the EXP function was used in the design of a number of other filters including the fuzzy vector median filter \cite{Plataniotis99}, fuzzy vector median-rational hybrid filter \cite{Khriji02}, kernel vector median filter \cite{Smolka03a}, fast adaptive noise reduction filter \cite{Smolka03b}, and self-adaptive noise reduction filter \cite{Smolka02}.

\begin{figure}[!ht]
\centering
\includegraphics[width=0.5\columnwidth,draft=false]{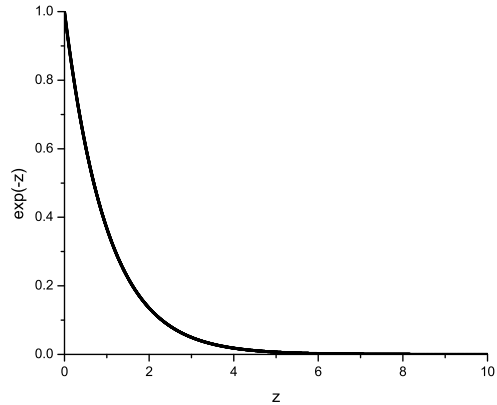}
\caption { \label{fig_exp} Function exp(-z) in the interval $[0,10]$ }
\end{figure}

The argument of the EXP function in (\ref{equ_amnf}) depends on the $\kappa$ value, and the size and contents of a particular window. However, to obtain an accurate approximation, this argument needs to be constrained to a preferably small interval. Fortunately, for most practical purposes, we can set a cutoff point at $T = 10.0$ ($e^{- T} = $ 4.539993e-05) and return 0 for arguments outside the interval $[0,T]$. Figure \ref{fig_exp} shows a plot of the function in this interval.

Table \ref{tab_exp_poly} shows the coefficients of the minimax polynomials of various degrees. Here, $p$ and $\varepsilon$ represent the degree of the polynomial and the error of minimax approximation, respectively. It can be seen that the error values are relatively high, and as the approximation degree is increased, the accuracy doesn't improve significantly. This suggests that rational functions might be better suited for this approximation. Table \ref{tab_exp_rat} lists the coefficients of a minimax rational that approximates the EXP function with an error of $\varepsilon = $ 2.227050e-06.

\begin{table}
\centering
\scriptsize
{
\caption{ \label{tab_exp_poly} Minimax polynomials for the EXP function }
\begin{tabular}{ c|c|c|c|c|c|c }
\hline
$p$ & $\varepsilon$ & $a_0$ & $a_1$ & $a_2$ & $a_3$ & $a_4$\\
\hline
\hline
2 & 1.785517e-01 & 8.214528e-01 & -3.186948e-01 & 2.544088e-02 &  & \\
\hline
3 & 8.259345e-02 & 9.174126e-01 & -5.631179e-01 & 1.015041e-01 & -5.519183e-03 & \\
\hline
4 & 3.337085e-02 & 9.666313e-01 & -7.620584e-01 & 2.145386e-01 & -2.509526e-02 & 1.032877e-03 \\
\hline
\end{tabular}
}
\end{table}

\begin{table}
\centering
\scriptsize
{
\caption{ \label{tab_exp_rat} Minimax rational for the EXP function }
\begin{tabular}{ c|c|c|c|c|c }
\hline
Term & $a_0$ & $a_1$ & $a_2$ & $a_3$ & $a_4$\\
\hline
\hline
Numerator & 3.206619e-02 & -1.195191e-02 & 1.756974e-03 & -1.199261e-04 & 3.182685e-06\\
\hline
Denominator & 3.206627e-02 & 2.011147e-02 & 5.853684e-03 & 9.780143e-04 & 1.251598e-04\\
\hline
\end{tabular}
}
\end{table}

\subsection{Entropy Vector Filters}
\label{sec_evmf}

Entropy vector median filter (EVMF) introduced in \cite{Lukac03a} adaptively switches between the identity operation and a noise filtering mode to improve signal-detail preserving characteristics of standard filters such as the vector median filter, which performs fixed amount of smoothing in all pixel locations. Noise filtering is performed only in pixel locations which are identified as noisy by a switching operator. This is realized by comparing an adaptive threshold $\beta_C$ expressed in the form of normalized entropy to a measure of normalized local contrast $P_C$ as follows:
\begin{equation}
\label{equ_evf}
\begin{array}{l}
  \mathbf{y}(r,c) = \left\{ {\begin{array}{*{20}c}
   {\mathop {\rm argmin}\limits_{\mathbf{x}_i \in W(r,c)} \left( {\sum\limits_{j = 1}^n {\left\| {\mathbf{x}_i  - \mathbf{x}_j } \right\|_2 } } \right)} \hfill & {P_C  > \beta _C } \hfill  \\
   {\mathbf{x}(r,c)} \hfill & \mbox{otherwise} \hfill  \\

 \end{array} } \right. \hfill \\
  P_i  = \frac{{\left\| {\mathbf{x}_i  - \mathbf{\bar x}} \right\|_2 }}
{{\sum\nolimits_{j = 1}^n {\left\| {\mathbf{x}_j  - \mathbf{\bar x}} \right\|} _2 }};\quad \beta _i  = \frac{{ - P_i \log P_i }}
{{ - \sum\nolimits_{j = 1}^n {P_j \log P_j } }} \hfill \\
\end{array}
\end{equation}
where $C = (n + 1)/2$ and $\bar{\bf x}$ denote the linear index of the center pixel (see Figure \ref{fig_window}) and the mean vector inside $W(r,c)$, i.e.\ $\bar{\bf x} = \frac{1}{n}\sum\nolimits_{i = 1}^n {{\bf x}_i}$, respectively.
\par
Note that within the so-called generalized entropy vector filter (EVF) class \cite{Lukac03b}, new filters can be designed by replacing the Euclidean distance function in (\ref{equ_evf}) with some other distance or similarity measure.
\par
\begin{figure}[!ht]
\centering
\includegraphics[width=0.5\columnwidth,draft=false]{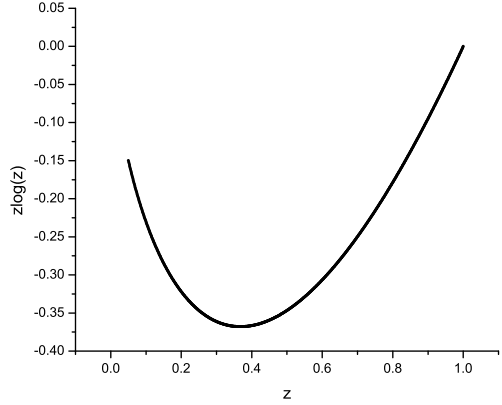}
\caption { \label{fig_ent} Function zlog(z) in the interval $[0.05,1]$ }
\end{figure}

The computational requirements of EVFs can be reduced by speeding up the entropy (ENT) function, whose argument falls into the interval $[0,1]$. Although, in theory, as the argument approaches $0$, the function value approaches $0$, in practice, this doesn't hold as the value of the logarithm function approaches negative infinity. Therefore, as in the case of the EXP function, we set a cutoff point at $T=0.05$ and return $0$ for arguments less than $T$. Figure \ref{fig_ent} shows a plot of the function in the interval $[0.05,1]$. It can be seen that  this function is highly nonpolynomial \cite{Muller06}, i.e.\ its derivatives are infinite at $z=0$, and therefore using rational functions is more appropriate. Table \ref{tab_ent_rat} lists the coefficients of a minimax rational that approximates the ENT function with an error of $\varepsilon = $ 7.342477e-07.

\begin{table}
\centering
\scriptsize
{
\caption{ \label{tab_ent_rat} Minimax rational for the ENT function }
\begin{tabular}{ c|c|c|c|c|c }
\hline
Term & $a_0$ & $a_1$ & $a_2$ & $a_3$ & $a_4$\\
\hline
\hline
Numerator & -1.519742e-04 & -6.835769e-02 & -8.856923e-01 & -5.369609e-01 & 1.491165\\
\hline
Denominator & 1.532270e-02 & 3.987796e-01 & 1.461793 & 6.827004e-01 & -4.469776e-02\\
\hline
\end{tabular}
}
\end{table}

\section{Experimental Results}
\label{sec_exp}

In order to evaluate the performance and robustness of the presented approximations, a set of 100 high quality RGB images was collected from the Internet. The  set included images of people, animals, plants, buildings, aerial maps, man-made objects, natural scenery, paintings, sketches, as well as scientific, biomedical, synthetic, and test images commonly used in the color image processing literature.
\par
The corruption in the test images was simulated using three noise models \cite{Viero94}: uncorrelated impulsive noise model, correlated impulsive noise model, and mixed noise model (Gaussian Noise + Correlated Impulsive Noise):
\begin{equation}
\label{eq_noise}
\begin{array}{l}
\mbox{Uncorrelated Impulsive Noise} \hfill \\
  {\bf x} = \left\{ {x_1 ,x_2 ,x_3} \right\} \hfill \\
  x_k  = \left\{ \begin{array}{l}
  o_k \,\,\,{\rm with}\,\,{\rm probability}\,\,1 - \varphi_k \,, \hfill \\
  r_k \,\,\,{\rm with}\,\,{\rm probability}\,\,\varphi_k \hfill \\
\end{array}  \right. \hfill \\

\mbox{Correlated Impulsive Noise} \hfill \\
{\bf x} = \left\{ \begin{array}{l}
 {\bf o}\quad \quad \quad \qquad\, {\rm with}\,\,{\rm probability}\,\,1 - \varphi \,, \\
 \left\{ {r_1 ,o_2 ,o_3 } \right\}\quad {\rm with}\,\,{\rm probability}\,\,\varphi _1  \cdot \varphi , \\
 \left\{ {o_1 ,r_2 ,o_3 } \right\}\quad {\rm with}\,\,{\rm probability}\,\,\varphi _2  \cdot \varphi , \\
 \left\{ {o_1 ,o_2 ,r_3 } \right\}\quad {\rm with}\,\,{\rm probability}\,\,\varphi _3  \cdot \varphi , \\
 \left\{ {r_1 ,r_2 ,r_3 } \right\}\quad {\rm with}\,\,{\rm probability}\,\,\left( {1 - (\varphi _1  + \varphi _2  + \varphi _3 )} \right) \cdot \varphi  \\
 \end{array} \right.
\end{array}
\end{equation}
where ${\bf o} = \left\{ {o_1,o_2,o_3} \right\}$ and ${\bf x} = \left\{ {x_1,x_2,x_3} \right\}$ represent the original and noisy color vectors, respectively, ${\bf r} = \left\{ {r_1,r_2,r_3} \right\}$ is a random vector that represents the impulsive noise, $\varphi $ is the sample corruption probability, and $\varphi _1$, $\varphi _2$, and $\varphi _3$ are the corruption probabilities for the red, green, and blue channels, respectively. In the experiments, the channel corruption probabilities were set to $0.25$.
\par
Filtering performance was evaluated by three effectiveness criteria \cite{Celebi07a}:
\begin{enumerate}
	\item Mean Absolute Error: $ \mbox{MAE}\left( {{\bf X},{\bf Y}} \right) = \frac{1}{{3MN}}\sum\limits_{r = 1}^M {\sum\limits_{c = 1}^N {{\parallel {\bf x}(r,c) - {\bf y}(r,c) \parallel}_1} } $\\
	where, ${\bf X}$ and ${\bf Y}$ denote respectively the $M \times N$ original and filtered images in the RGB color space. MAE measures the detail preservation capability of a filter.
	\item Mean Squared Error: $ \mbox{MSE}\left( {{\bf X},{\bf Y}} \right) = \frac{1}{{3MN}}\sum\limits_{r = 1}^M {\sum\limits_{c = 1}^N {\parallel {\bf x}(r,c) - {\bf y}(r,c) \parallel}_2^2} $\\
	MSE measures the noise suppression capability of a filter.
	\item Normalized Color Difference: $ \mbox{NCD}\left( {{\bf X},{\bf Y}} \right) = \frac{{\sum\limits_{r = 1}^M {\sum\limits_{c = 1}^N {{\parallel {\bf x}^{Lab}(r,c) - {\bf y}^{Lab}(r,c) \parallel}_2} } }}{{\sum\limits_{r = 1}^M {\sum\limits_{c = 1}^N {\left\| {{\bf x}^{Lab}(r,c)} \right\|_2 } } }} $\\
	where, ${\bf x}^{Lab}(r,c)$ and ${\bf y}^{Lab}(r,c)$ denote the CIEL*a*b* coordinates \cite{Plataniotis00} of the pixel $(r,c)$ in the original and filtered images, respectively. NCD measures the color preservation capability of a filter.
\end{enumerate}
The efficiency of a filter was measured by execution time in seconds (Programming Language: C, Compiler: gcc 3.4.4, CPU: Intel Pentium D 2.66Ghz).
\par
Table \ref{tab_stats} shows the performance statistics for the three noise models. The test images were first corrupted using one of the noise models and then filtered using the exact and approximate versions of each filter. In the 'Mean' column, negative values and positive values for the MAE, MSE, and NCD indicate the percentage of filtering quality degradation and improvement, respectively. For example, for 10\% correlated impulsive noise, with respect to the MAE criterion, the approximate version of BVDF performs on the average 0.926\% worse than the exact version, whereas with respect to the MSE criterion, the former performs 0.171\% better than the latter. On the other hand, for the execution time criterion, positive values indicate reduction in filtering time due to the use of the presented approximations. For example, the approximate version of BVDF is on the average 1371\% (or 13.71 times) faster than the exact version.

\begin{table}
\centering
\scriptsize
{
\caption{ \label{tab_stats} Performance statistics at 10\% noise level}
\begin{tabular}{ c|c|c|c|c|c|c|c }
\hline
\multicolumn{2}{c|}{} & \multicolumn{2}{|c|}{Uncorrelated
Impulsive} & \multicolumn{2}{|c|}{Correlated Impulsive} & \multicolumn{2}{|c}{Mixed}\\
\hline
Filter & Measure & Mean(\%) & Stdev(\%) & Mean(\%) & Stdev(\%) & Mean(\%) & Stdev(\%)\\
\hline
\hline
\multirow{4}{*}{BVDF} & MAE & -1.000 & 1.648 & -0.926 & 1.558 & -0.022 & 0.310\\
 & MSE & -0.571 & 1.924 & 0.171 & 2.426 & -0.131 & 1.103\\
 & NCD & -0.783 & 1.283 & -0.759 & 1.154 & 0.007 & 0.193\\
 & Time & 1381.783 & 16.576 & 1371.314 & 16.479 & 1408.964 & 11.964\\
\hline
\multirow{4}{*}{AMNFE} & MAE & -0.025 & 0.137 & -0.016 & 0.136 & -0.001 & 0.137\\
 & MSE & -0.070 & 0.526 & -0.064 & 0.754 & -0.006 & 0.317\\
 & NCD & -0.020 & 0.177 & -0.031 & 0.195 & -0.004 & 0.176\\
 & Time & 142.700 & 0.225 & 143.496 & 0.214 & 140.814 & 0.187\\
\hline
\multirow{4}{*}{EVMF} & MAE & -0.141 & 0.524 & -0.108 & 0.489 & 0.000 & 0.133\\
 & MSE & -0.376 & 1.732 & -0.203 & 1.496 & 0.013 & 0.241\\
 & NCD & -0.147 & 0.837 & -0.149 & 0.625 & -0.004 & 0.138\\
 & Time & 236.582 & 0.672 & 236.105 & 0.646 & 215.000 & 0.449\\
\hline
\end{tabular}
}
\end{table}


It can be seen that in most cases the exact filters slightly outperform their respective approximate versions. This was expected since the approximate filters necessarily involve small amounts of computational error. Nevertheless, the difference between the approximate and exact versions for each filter is negligible for most practical purposes, which demonstrates the accuracy of the presented approximations. In addition, the low standard deviation values indicate the robustness of the approximations.
\par
The discrepancies in the speed up factors for the three filters can be attributed to the relative computational cost of the elementary functions involved. In other words, the speed up in BVDF is much greater than the other two filters because the ARCCOS function is computationally much more expensive than the EXP and ENT functions.
\par
Since the filters presented in Section \ref{sec_impl} are primarily intended for the removal of impulsive noise, we conducted further experiments with the most commonly used impulsive noise model \cite{Lukac06a,Celebi07a}, i.e. the correlated impulsive noise model \cite{Viero94}. Table \ref{tab_stats_more} shows the performance statistics at 20\%, 30\%, and 40\% noise levels. It can be seen that the performance of the approximate filters does not change significantly as the noise level is increased.

\begin{table}
\centering
\scriptsize
{
\caption{ \label{tab_stats_more} Performance statistics at higher noise levels}
\begin{tabular}{ c|c|c|c|c|c|c|c }
\hline
\multicolumn{2}{c|}{} & \multicolumn{2}{|c|}{20\%} & \multicolumn{2}{|c|}{30\%} & \multicolumn{2}{|c}{40\%}\\
\hline
Filter & Measure & Mean(\%) & Stdev(\%) & Mean(\%) & Stdev(\%) & Mean(\%) & Stdev(\%)\\
\hline
\hline
\multirow{4}{*}{BVDF} & MAE & -0.753 & 1.311 & -0.626 & 1.070 & -0.419 & 0.861\\
 & MSE & 0.193 & 2.994 & 0.027 & 2.630 & -0.073 & 1.855\\
 & NCD & -0.607 & 1.016 & -0.530 & 0.881 & -0.424 & 0.802\\
 & Time & 1342.074 & 15.480 & 1328.939 & 14.784 & 1301.070 & 14.026\\
\hline
\multirow{4}{*}{AMNFE} & MAE & -0.032 & 0.231 & 0.030 & 0.342 & -0.001 & 0.315\\
 & MSE & -0.125 & 0.945 & 0.021 & 1.172 & 0.006 & 0.812\\
 & NCD & -0.045 & 0.349 & 0.034 & 0.478 & 0.027 & 0.395\\
 & Time & 143.460 & 0.241 & 145.513 & 0.226 & 145.362 & 0.249\\
\hline
\multirow{4}{*}{EVMF} & MAE & -0.030 & 0.676 & 0.013 & 0.534 & 0.067 & 0.476\\
 & MSE & -0.056 & 1.539 & 0.046 & 1.105 & 0.065 & 0.867\\
 & NCD & -0.004 & 0.829 & 0.026 & 0.633 & 0.088 & 0.547\\
 & Time & 228.346 & 0.551 & 222.356 & 0.431 & 216.354 & 0.423\\
\hline
\end{tabular}
}
\end{table}

\begin{figure}[!ht]
\centering
 \subfigure[Original ($512 \times 512$)]{\label{lenna_a}\includegraphics[width=0.26\columnwidth]{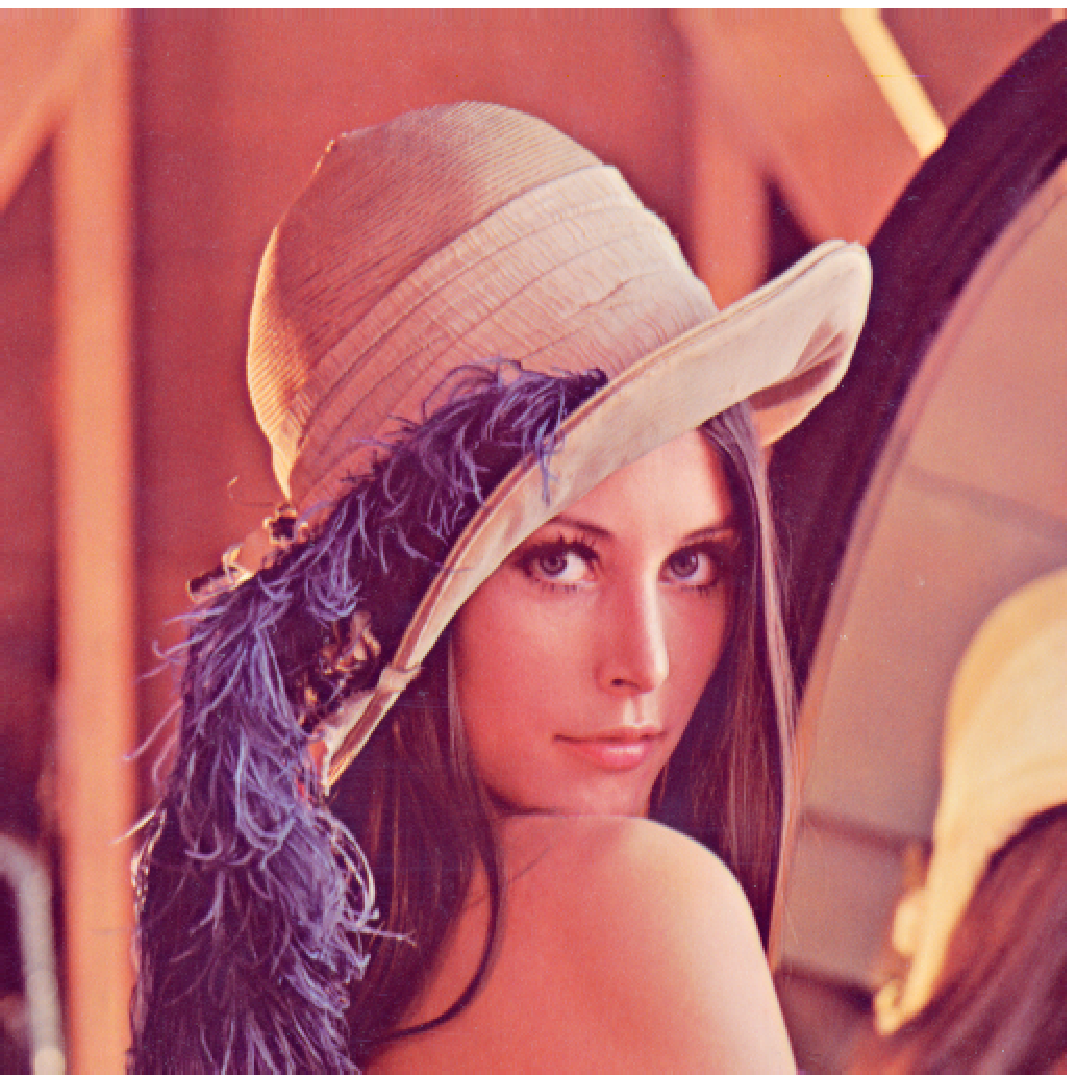}}
 \hspace{0.02in}
 \subfigure[10\% noisy (MAE 6.373, MSE 987.418, Time 0)]
 {\label{lenna_b}\includegraphics[width=0.26\columnwidth]{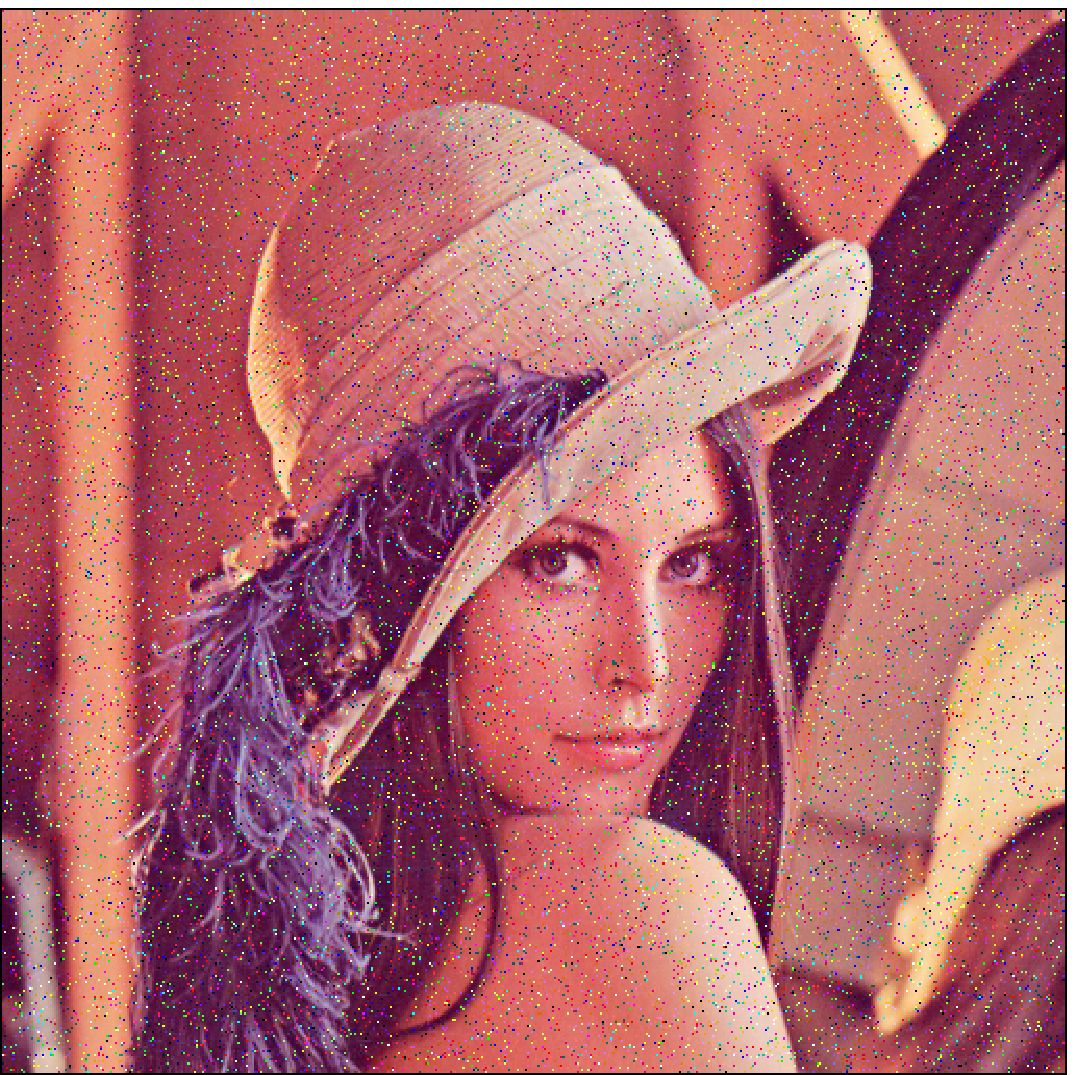}}
 \hspace{0.02in}
 \subfigure[BVDF exact (MAE 3.936, MSE 43.571, Time 10.360)]{\label{lenna_c}\includegraphics[width=0.26\columnwidth]{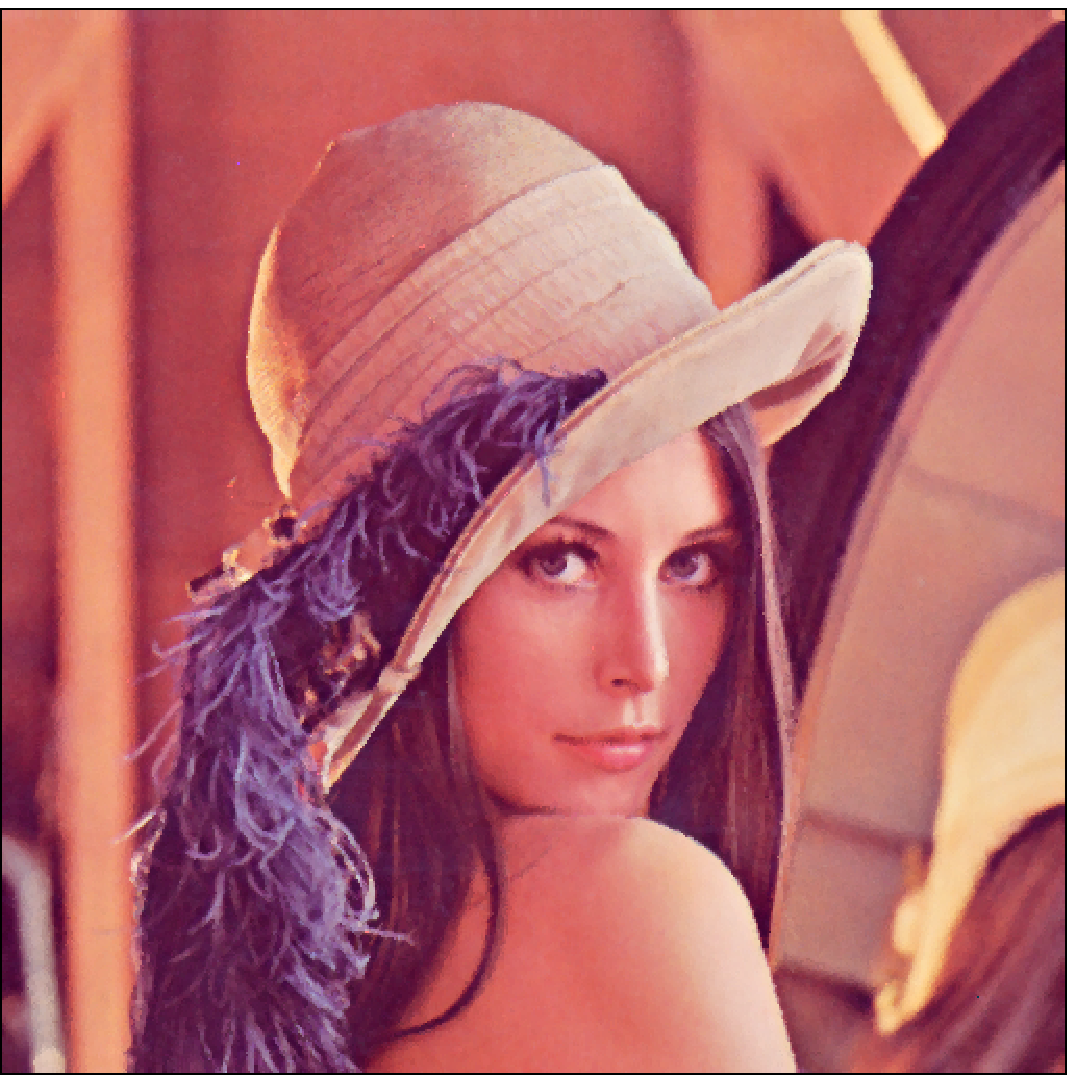}}
 \\
 \subfigure[BVDF approx (MAE 3.936, MSE 43.558, Time 0.688)]{\label{lenna_d}\includegraphics[width=0.26\columnwidth]{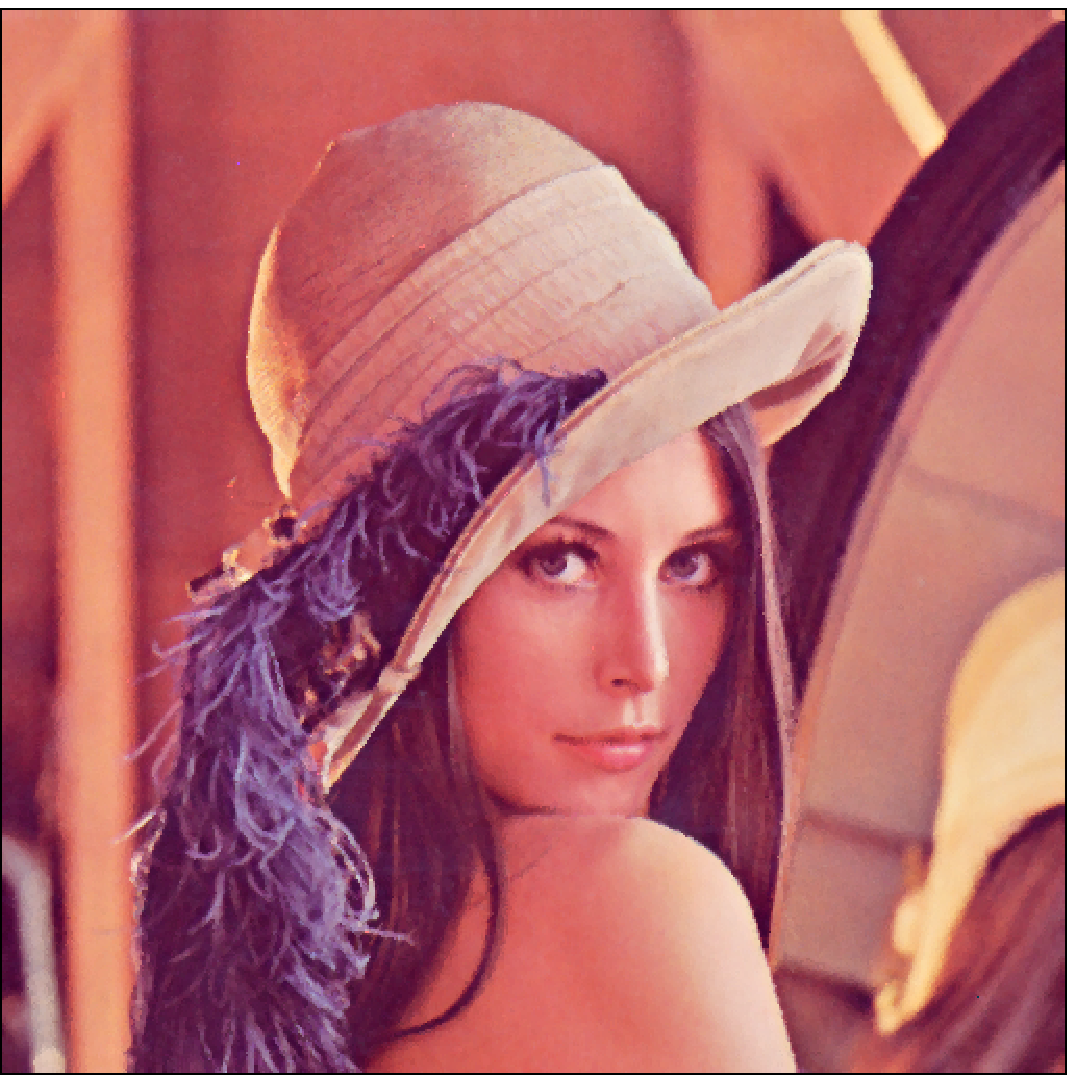}}
 \hspace{0.02in}
 \subfigure[AMNFE exact (MAE 3.471, MSE 30.303, Time 0.594)]{\label{lenna_e}\includegraphics[width=0.26\columnwidth]{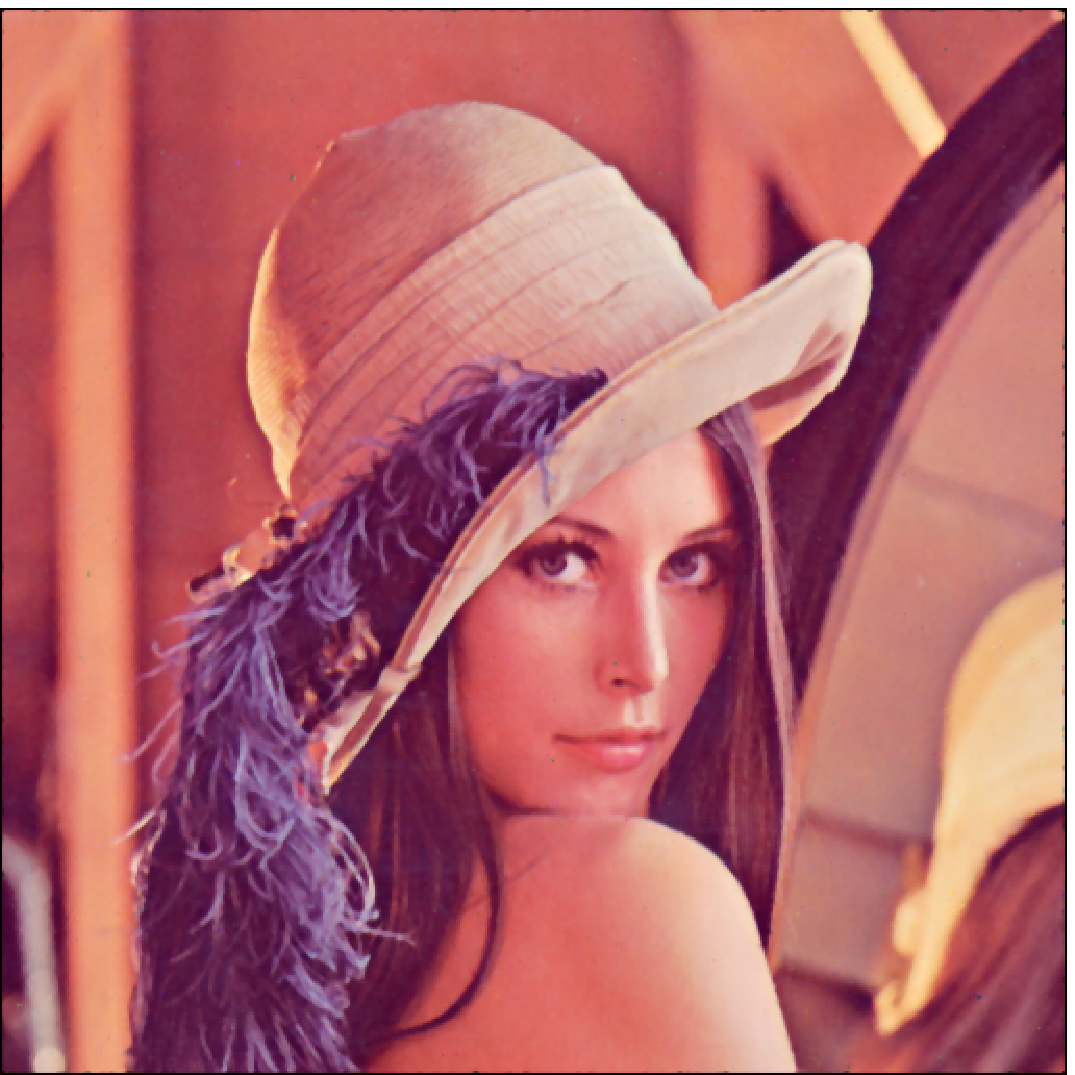}}
 \hspace{0.02in}
 \subfigure[AMNFE approx (MAE 3.471, MSE 30.303, Time 0.422)]{\label{lenna_f}\includegraphics[width=0.26\columnwidth]{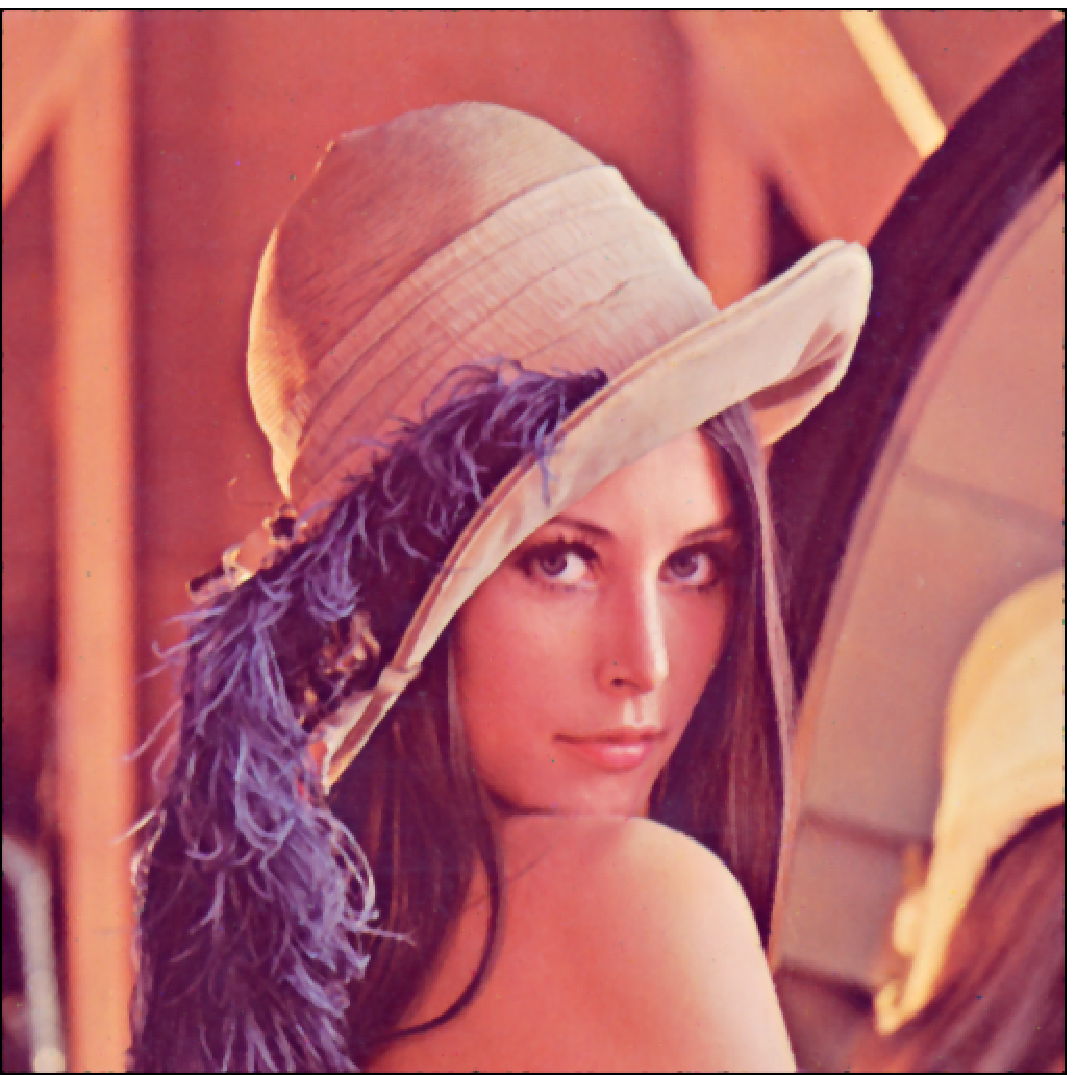}}
 \\
 \subfigure[EVMF exact (MAE 1.139, MSE 25.898, Time 0.594)]{\label{lenna_g}\includegraphics[width=0.26\columnwidth]{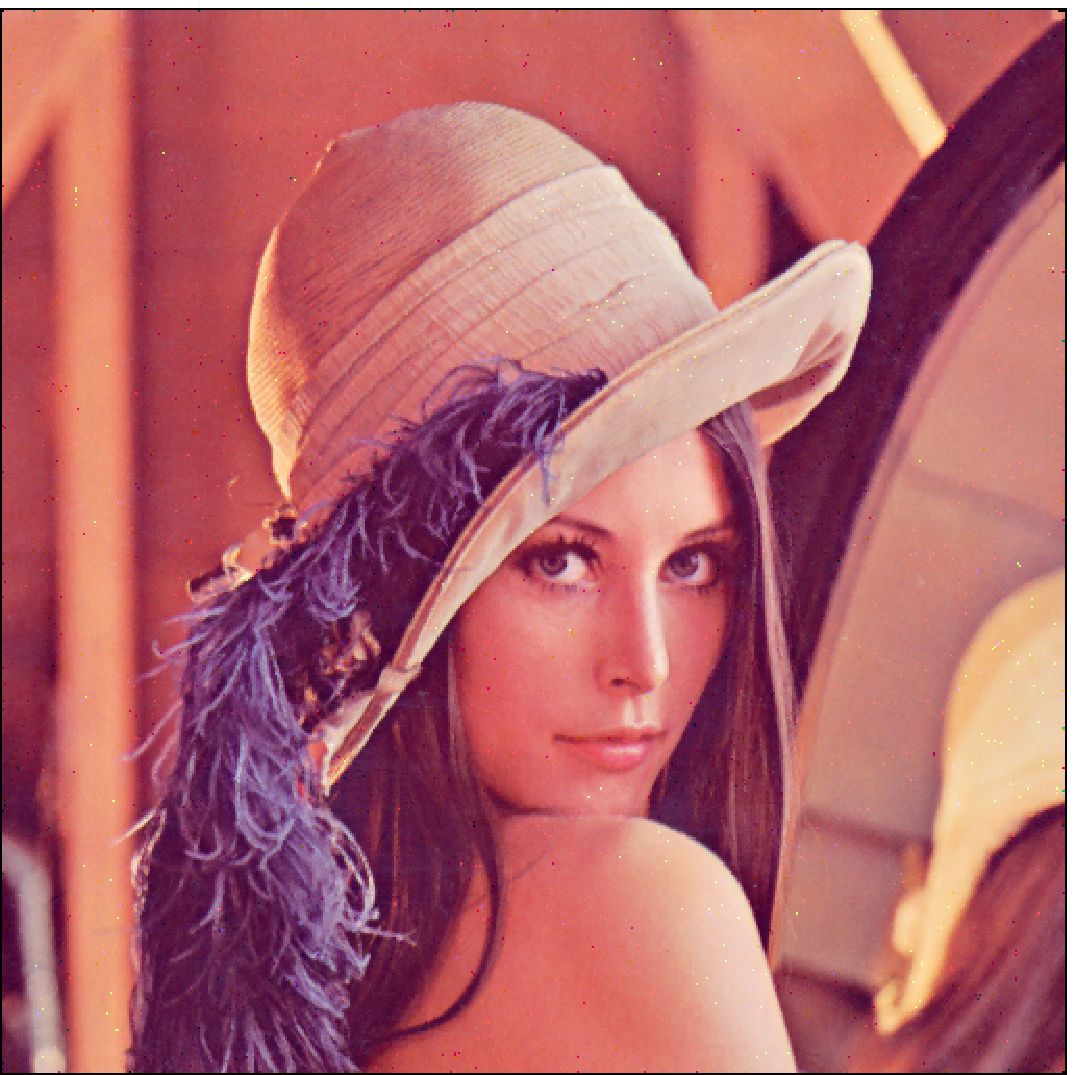}}
 \hspace{0.02in}
 \subfigure[EVMF approx (MAE 1.138, MSE 25.819, Time 0.250)]{\label{lenna_h}\includegraphics[width=0.26\columnwidth]{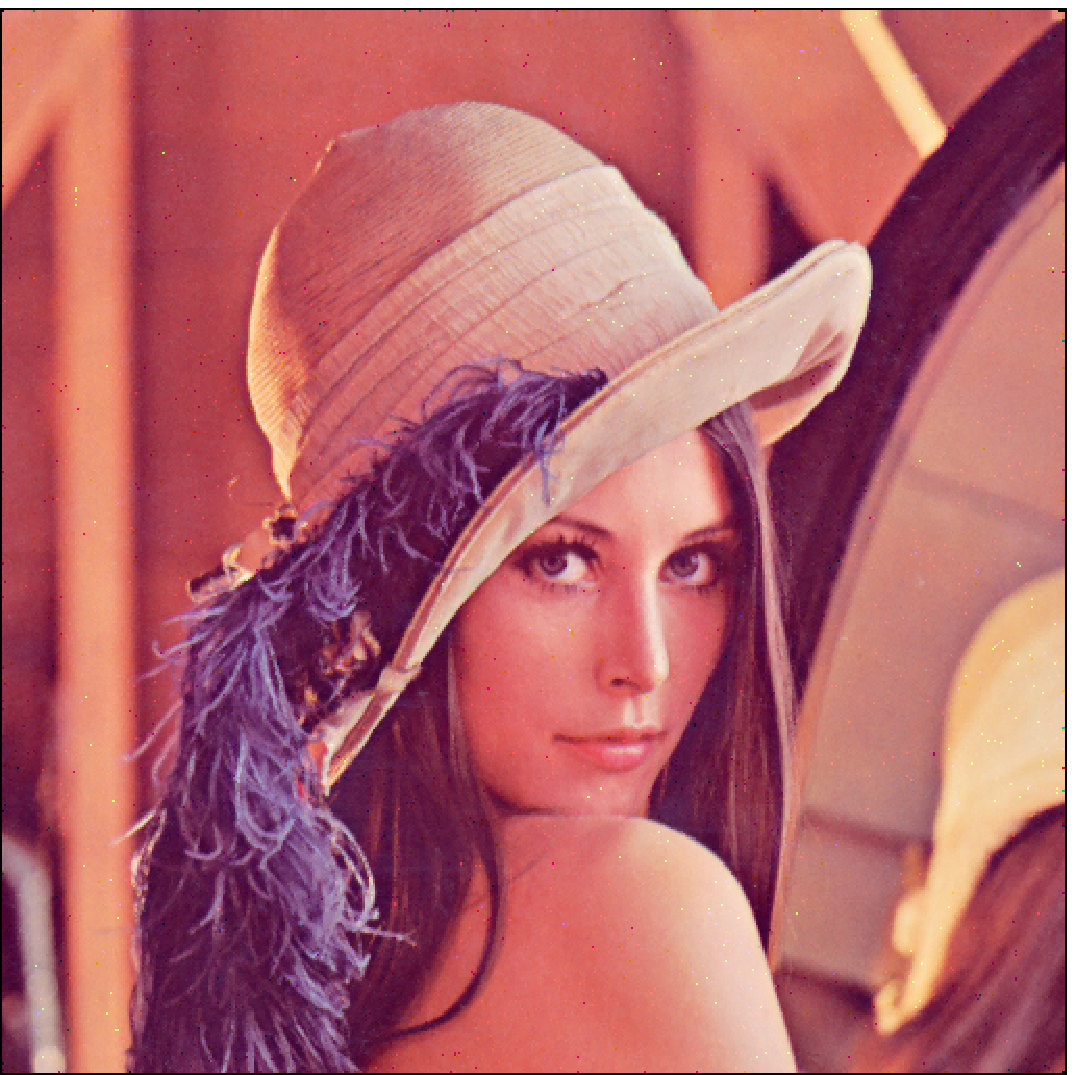}}
 \caption{Comparison of the exact and approximate filters on the Lenna image}
 \label{fig_lenna}
\end{figure}

Figure \ref{fig_lenna} compares the exact and approximate  versions of each filter on the Lenna image. It can be seen that the presented approximations achieve substantial computational savings without introducing any perceivable artifacts on the filtering results. In addition, the MAE and MSE values indicate that the filtering effectiveness of the exact and approximate filters are virtually the same.

\section{Conclusions}
\label{sec_conc}

In this article, we proposed a novel approach to speed up popular vector filters using minimax approximations. Advantages of this approach include ease of implementation, extremely good accuracy, and high computational speed. The presented approach can be adapted to other noise removal filters that involve computationally expensive mathematical functions. Finally, the given approximations have applications that go beyond color image filtering including computer graphics and computational geometry.

\section*{Acknowledgments}
This publication was made possible by a grant from The Louisiana Board of Regents (LEQSF2008-11-RD-A-12).

\end{document}